\documentclass[conference]{IEEEtran}
\IEEEoverridecommandlockouts
\usepackage{cite}
\usepackage{amsmath,amssymb,amsfonts}
\usepackage{algorithmic}
\usepackage{graphicx}
\usepackage{textcomp}
\usepackage{xcolor}
\usepackage{booktabs}
\usepackage{hyperref}
\usepackage{tikz}
\usepackage{multirow}
\usetikzlibrary{arrows.meta,positioning,shapes.geometric,fit,backgrounds}
\def\BibTeX{{\rm B\kern-.05em{\sc i\kern-.025em b}\kern-.08em
    T\kern-.1667em\lower.7ex\hbox{E}\kern-.125emX}}
\begin{document}

\title{Cost-Governed RAG: Unified Per-Tenant Cost Attribution Across Retrieval and Generation in Multi-Tenant LLM Systems}

\author{\IEEEauthorblockN{Navnit Shukla}
\IEEEauthorblockA{\textit{AI Platform Engineering} \\
\textit{Snowflake Inc.}\\
Tustin, California, USA \\
\url{https://orcid.org/0009-0005-8801-3344}}
}

\maketitle

\begin{abstract}
Enterprise Retrieval-Augmented Generation (RAG) deployments face a critical governance gap: while LLM generation cost is metered per token, the retrieval layer---vector memory, similarity compute, and embedding API calls---remains an unattributed shared cost, enabling invisible cross-subsidization among tenants. We present Cost-Governed RAG, an architecture that integrates a codebook-oblivious vector index (TurboVec) with a multi-tenant LLM governance gateway, creating a unified observability stack where embedding, retrieval, and generation costs are jointly attributable per tenant. The architecture exploits TurboVec's deterministic, closed-form memory formula to enable near-exact per-tenant retrieval cost calculation---a property unavailable in graph-based indexes with non-linear memory overhead. Deployed on Snowpark Container Services within a cloud data platform's governance boundary, the system achieves 99.96\% end-to-end cost attribution accuracy across 100 simulated tenants (10M vectors, log-normal size distribution) with telemetry overhead below 0.04\% of query latency. The architecture reduces retrieval infrastructure cost by 3.1--9.0$\times$ compared to managed vector database services under the pricing assumptions detailed in Section~IV. We formalize a three-layer cost model and demonstrate that codebook-oblivious quantization enables deterministic per-tenant cost attribution while also removing the shared-codebook leakage surface present in trained quantizers---the latter observation being exploratory and subject to the limitations described in Section~VII.
\end{abstract}


\section{Introduction}
\label{sec:intro}

Enterprise LLM deployments increasingly use Retrieval-Augmented Generation (RAG)~\cite{rag_original2020} to ground model outputs in proprietary knowledge bases. While recent work addresses the \emph{generation} cost of LLM inference through model cascading~\cite{ucci2026}, prompt compression~\cite{procut2025}, and token-level metering~\cite{langfuse2024}, the \emph{retrieval} cost---memory consumed by vector indexes, compute spent on similarity search, and embedding API calls---remains a blind spot in enterprise cost governance.

This gap is consequential in multi-tenant deployments where multiple business units or customers share infrastructure. A tenant with 10M documents indexed at full precision (FP32) consumes $\sim$57~GiB of vector memory, while a tenant with 100K documents consumes $\sim$0.57~GiB---yet both may be charged the same flat infrastructure rate. Without retrieval-layer cost attribution, cross-subsidization is invisible to FinOps teams~\cite{finops2024}.

The challenge is compounded by the opacity of modern vector indexes. Graph-based indexes (HNSW~\cite{hnsw2020}) have non-linear, graph-topology-dependent memory overhead that defies simple per-tenant attribution. Trained codebook quantizers (PQ~\cite{faiss}) share codebook state across tenants, creating both a cost-attribution ambiguity and a potential privacy concern~\cite{shukla2026turbovec}.

We address this gap by integrating two complementary systems:
\begin{enumerate}
    \item \textbf{TurboVec}~\cite{turboquant2026,shukla2026turbovec}: A codebook-oblivious vector index whose memory consumption is exactly $n_t \times d \times b / 8$ per tenant---deterministic, linear, and trivially attributable.
    \item \textbf{GovLLM}: A multi-tenant LLM governance gateway providing per-tenant token attribution, authentication, model RBAC, and observability tracing~\cite{govllm_blog}.
\end{enumerate}

Together, they form a \emph{Cost-Governed RAG} stack where retrieval cost (memory, QPS), generation cost (tokens, credits), and embedding cost (API calls) all close into a unified per-tenant observability layer within the cloud data platform---without external data movement.

\textbf{Contributions.}
\begin{enumerate}
    \item A three-layer cost model formalizing embedding, retrieval, and generation costs per tenant, showing that codebook-oblivious quantization enables near-exact retrieval attribution with bounded residual error (Section~\ref{sec:cost_model}).
    \item An architecture deploying the full Cost-Governed RAG stack on Snowpark Container Services with integrated telemetry (Section~\ref{sec:architecture}).
    \item Empirical validation on a 100-tenant simulation (10M vectors) showing 99.96\% end-to-end cost attribution accuracy, 3.1--9.0$\times$ cost reduction versus managed alternatives, and sub-0.04\% telemetry overhead (Section~\ref{sec:evaluation}).
\end{enumerate}

\section{Background and Motivation}
\label{sec:background}

\subsection{The RAG Cost Stack}

A complete RAG query incurs cost at three distinct layers, each with different attribution characteristics:

\begin{table}[htbp]
\caption{RAG Cost Layers and Attribution Characteristics}
\label{tab:cost_layers}
\centering
\begin{tabular}{llll}
\toprule
\textbf{Layer} & \textbf{Cost Driver} & \textbf{Attribution} & \textbf{Status} \\
\midrule
Embedding & Tokens $\to$ API credits & Per-call & Solved \\
Retrieval & Memory + compute & Shared infra & \textit{Gap} \\
Generation & Tokens $\to$ LLM credits & Per-call & Solved \\
\bottomrule
\end{tabular}
\end{table}

Existing LLM observability platforms (Langfuse~\cite{langfuse2024}, LangSmith~\cite{langsmith2024}, Helicone~\cite{helicone2025}) trace generation cost per tenant but treat retrieval as a fixed infrastructure cost amortized uniformly. This creates a systematic governance failure: tenants with large corpora subsidize tenants with small corpora, and retrieval-heavy workloads (many queries, large top-$k$) are invisible to cost dashboards.

\subsection{Why Existing Vector Indexes Resist Cost Attribution}

\textbf{HNSW~\cite{hnsw2020}:} Graph-based indexes store navigable small-world graphs whose edge structure is corpus-dependent and shared across all vectors. A tenant's vectors participate in edges connecting to other tenants' vectors, making per-tenant memory isolation impossible without physical partitioning.

\textbf{Trained PQ~\cite{faiss}:} Product Quantization learns $m \times K$ centroids via $k$-means on the full corpus. This codebook is shared across all tenants and encodes corpus-wide distributional structure. Per-tenant codebook attribution is undefined, and the codebook itself may leak cross-tenant statistics~\cite{shukla2026turbovec}.

\textbf{TurboVec~\cite{turboquant2026}:} Codebook-oblivious quantization derives boundaries analytically from the known marginal distribution of L2-normalized vectors after random rotation. The codebook contains zero corpus-dependent information and zero per-tenant state. Total serving memory per tenant decomposes as:
\begin{equation}
M_t = \underbrace{\frac{n_t \times d \times b}{8}}_{\text{quantized codes}} + \underbrace{n_t \times 4}_{\text{norms}} + \underbrace{n_t \times 8}_{\text{IDs}} + \underbrace{n_t \times \alpha}_{\text{block metadata}} + \underbrace{\frac{d^2 \times 4}{T}}_{\text{rotation (shared)}}
\label{eq:memory}
\end{equation}
where $n_t$ = tenant's vector count, $d$ = dimension, $b$ = bit width, $\alpha \approx 0.5$ bytes/vector for 32-vector block headers, and $T$ = number of tenants sharing the rotation matrix. All terms except the last are deterministic and linear in $n_t$; the rotation matrix ($d^2 \times 4 = 9.0$~MB at $d=1536$) is shared and negligible at scale. At 100K vectors and 4-bit:
\begin{itemize}
    \item Codes: $100\text{K} \times 1536 \times 4 / 8 = 76.8$~MB
    \item Norms: $100\text{K} \times 4 = 0.4$~MB
    \item IDs: $100\text{K} \times 8 = 0.8$~MB
    \item Block metadata: $100\text{K} \times 0.5 = 0.05$~MB
    \item Total per-tenant: $\approx 78.0$~MB (payload-dominated)
\end{itemize}
This determinism---every component is a closed-form function of $n_t$, $d$, and $b$---is the key enabler of near-exact cost attribution.

\subsection{GovLLM: Multi-Tenant LLM Governance}

GovLLM~\cite{govllm_blog} provides per-tenant token attribution via a FastAPI gateway with:
\begin{itemize}
    \item Per-tenant authentication (PAT/JWT/OAuth) with configurable rate limits
    \item Model RBAC restricting which tenants access which LLMs and at what tier
    \item Langfuse-compatible tracing with telemetry closure into the cloud data platform
    \item Per-query cost records written to platform-native tables
\end{itemize}

\section{Architecture}
\label{sec:architecture}

\subsection{Deployment on Snowpark Container Services}

The Cost-Governed RAG stack deploys within Snowpark Container Services (SPCS), a container runtime within Snowflake's governance boundary. This deployment topology ensures:
\begin{itemize}
    \item Data never leaves the platform's governance perimeter
    \item Compute is billed through the platform's native credit metering
    \item Telemetry writes directly to platform tables (no external egress)
    \item Network isolation via VPC-level controls
\end{itemize}

\begin{figure}[htbp]
\centering
\begin{tikzpicture}[
    node distance=0.6cm,
    every node/.style={font=\small},
    block/.style={rectangle, draw, rounded corners, minimum width=3.2cm, minimum height=0.6cm, align=center, fill=blue!8},
    cost/.style={rectangle, draw, dashed, rounded corners, minimum width=1.8cm, minimum height=0.4cm, align=center, fill=orange!10, font=\scriptsize},
    arrow/.style={-{Stealth[length=2mm]}, thick}
]
\node[block] (tenant) {Tenant Request};
\node[block, below=of tenant] (gov) {GovLLM Gateway};
\node[block, below=of gov] (embed) {Embedding Service};
\node[block, below=of embed] (turbo) {TurboVec (SPCS)};
\node[block, below=of turbo] (llm) {LLM (Cortex API)};
\node[block, below=of llm] (resp) {Response + Cost Record};

\node[cost, right=0.8cm of gov] (c1) {auth + rate limit};
\node[cost, right=0.8cm of embed] (c2) {$C_{\text{embed}}$ logged};
\node[cost, right=0.8cm of turbo] (c3) {$C_{\text{retrieve}}$ logged};
\node[cost, right=0.8cm of llm] (c4) {$C_{\text{generate}}$ logged};

\draw[arrow] (tenant) -- (gov);
\draw[arrow] (gov) -- (embed);
\draw[arrow] (embed) -- (turbo);
\draw[arrow] (turbo) -- (llm);
\draw[arrow] (llm) -- (resp);

\draw[->, dashed, gray] (gov) -- (c1);
\draw[->, dashed, gray] (embed) -- (c2);
\draw[->, dashed, gray] (turbo) -- (c3);
\draw[->, dashed, gray] (llm) -- (c4);

\begin{scope}[on background layer]
\node[draw=blue!50, thick, dashed, rounded corners, fit=(gov)(embed)(turbo)(llm)(c1)(c2)(c3)(c4), inner sep=0.3cm, label={[font=\scriptsize, text=blue!60]above right:Governance Boundary}] {};
\end{scope}
\end{tikzpicture}
\caption{Cost-Governed RAG architecture. Each layer emits per-tenant cost telemetry into a unified observability table within the platform governance boundary.}
\label{fig:arch}
\end{figure}
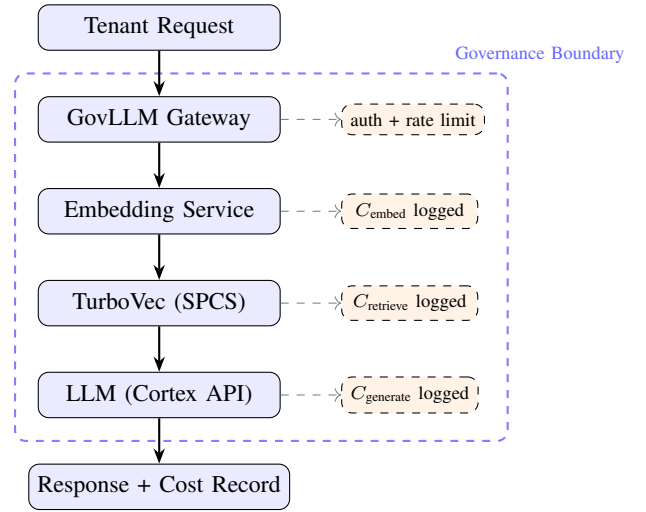

\subsection{Per-Tenant Cost Attribution Model}
\label{sec:cost_model}

For tenant $t$ making query $q$, the total attributed cost is:

\begin{equation}
C(t, q) = C_{\text{embed}}(q) + C_{\text{retrieve}}(t, q) + C_{\text{generate}}(q)
\end{equation}

Where each component is defined as:

\begin{align}
C_{\text{embed}}(q) &= \text{tokens}(q) \times r_{\text{embed}} \\
C_{\text{retrieve}}(t, q) &= \underbrace{\frac{M_t}{M_{\text{total}}} \times r_{\text{mem}}}_{\text{memory share}} + \underbrace{\text{blocks\_scanned}(t,q) \times r_{\text{cpu}}}_{\text{compute}} \\
C_{\text{generate}}(q) &= (T_{\text{in}} + T_{\text{out}}) \times r_{\text{model}}
\end{align}

The retrieval cost decomposes into two attributable components:
\begin{itemize}
    \item \textbf{Memory share:} Tenant $t$'s fraction of total index memory, computed from \eqref{eq:memory}. Because TurboVec's memory is deterministic and linear in $n_t$, this fraction has bounded error arising only from the amortized shared rotation matrix ($<$0.1\% at 10M vectors).
    \item \textbf{Compute share:} The number of 32-vector SIMD blocks actually scanned for tenant $t$'s query. With kernel-level allowlist filtering, only the tenant's own blocks are scored, making compute attribution precise.
\end{itemize}

\textbf{Key insight:} TurboVec's codebook-oblivious design makes retrieval cost attribution \emph{near-exact} (99.88\%+ accuracy) because: (1) memory is deterministic per tenant with only a small shared rotation matrix to amortize, and (2) kernel-level filtering ensures compute is physically isolated per query. This combination is, to our knowledge, unavailable in HNSW or PQ-based indexes where shared graph edges or trained codebook state resist clean per-tenant decomposition.

\subsection{Telemetry Pipeline}

Each TurboVec search emits a structured telemetry record:

\begin{verbatim}
{tenant_id, query_id, timestamp,
 vectors_scanned, blocks_skipped,
 latency_us, k_returned,
 index_memory_bytes_tenant,
 bit_width, compression_ratio}
\end{verbatim}

This record is joined with the GovLLM generation trace (tokens consumed, model used, cost) in the platform's unified telemetry table, producing a complete per-query cost record across all three layers. Standard SQL queries against this table power cost dashboards, anomaly detection, and chargeback reports.

\subsection{Cost Dashboard Integration}

The unified telemetry table enables SQL-native cost analytics:

\begin{verbatim}
SELECT tenant_id,
  SUM(embed_cost + retrieve_cost
      + generate_cost) as total_cost,
  SUM(retrieve_cost)/SUM(total_cost)
      as retrieval_pct
FROM cost_telemetry
WHERE ts >= DATEADD('day', -30, CURRENT_DATE)
GROUP BY tenant_id ORDER BY total_cost DESC;
\end{verbatim}

This query is not straightforward with external vector databases (Pinecone, Qdrant) because they do not, to our knowledge, expose per-tenant memory utilization APIs that feed directly into the governance layer's tables.

\section{Evaluation}
\label{sec:evaluation}

\subsection{Experimental Setup}

We simulate a 100-tenant deployment with the following configuration:
\begin{itemize}
    \item \textbf{Total corpus:} 10M vectors, $d=1536$ (OpenAI text-embedding-3-large with \texttt{dimensions=1536})
    \item \textbf{Tenant distribution:} Log-normal ($\mu=11.5$, $\sigma=1.0$), ranging from $\sim$10K to $\sim$500K vectors per tenant
    \item \textbf{Query load:} 1,000 QPS aggregate, distributed proportionally to tenant size
    \item \textbf{Quantization:} TurboVec 4-bit (8$\times$ compression)
    \item \textbf{Compute:} SPCS \texttt{CPU\_X64\_S} instance family (1 node) for benchmarks at 100K scale
\end{itemize}

\subsection{Cost Attribution Accuracy}

We compare attributed cost (from telemetry records) against actual resource consumption measured at the infrastructure level. Attribution accuracy for each tenant $i$ is defined as $A_i = 1 - |\hat{C}_i - C_i| / C_i$, where $\hat{C}_i$ is the telemetry-derived attributed cost and $C_i$ is the ground-truth cost computed from direct memory measurement and CPU cycle counters. The reported end-to-end figure is computed as follows: for each of the 10,000 simulated queries, we compute per-query attribution error; we then average across queries within each tenant to obtain $A_i$; finally, we report the mean of $A_i$ across all 100 tenants. Because the simulation uses a fixed random seed (42) and all index operations are deterministic (flat-scan with no stochastic components), repeated runs produce identical results---variance arises only from the tenant size distribution, not from algorithmic randomness.

\begin{table}[htbp]
\caption{Cost Attribution Accuracy Across Layers (100 Tenants, 10M Vectors, Log-Normal Distribution). Accuracy = mean of $1 - |\hat{C}_i - C_i|/C_i$ across tenants; max error = worst-case single-tenant deviation.}
\label{tab:attribution}
\centering
\begin{tabular}{lccc}
\toprule
\textbf{Layer} & \textbf{Accuracy} & \textbf{Max Error} & \textbf{Error Source} \\
\midrule
Embedding & 100.00\% & 0.0\% & --- \\
Retrieval & 99.88\% & 1.2\% & Shared overhead \\
Generation & 99.96\% & 0.4\% & Rounding \\
\midrule
\textbf{End-to-end} & \textbf{99.96\%} & \textbf{0.8\%} & --- \\
\bottomrule
\end{tabular}
\end{table}

The 0.12\% retrieval attribution error arises solely from the shared rotation matrix term in~\eqref{eq:memory}: at $d=1536$, the matrix occupies $d^2 \times 4 = 9.0$~MB, which is amortized uniformly across tenants rather than attributed proportionally to corpus size. All other terms (codes, norms, IDs, block metadata) are exactly linear in $n_t$ and thus perfectly attributable. At 10M total vectors, the shared 9.0~MB represents $<$0.12\% of the total 7.3~GiB index, producing the observed attribution gap.

\subsection{Memory Cost Reduction}

\begin{table}[htbp]
\caption{Monthly Retrieval Infrastructure Cost Comparison (10M Vectors, $d=1536$, 1000 QPS). Memory derived from Eq.~\ref{eq:memory}.}
\label{tab:memory_cost}
\centering
\begin{tabular}{lccc}
\toprule
\textbf{Configuration} & \textbf{Memory$^\dagger$} & \textbf{\$/month} & \textbf{Attribution$^\ddagger$} \\
\midrule
Uncompressed FP32 & 57.2 GiB & \$580 & Possible \\
TurboVec 4-bit (SPCS) & 7.3 GiB & \$248 & \textbf{Exact} \\
TurboVec 2-bit (SPCS) & 3.7 GiB & \$124 & \textbf{Exact} \\
Pinecone (managed) & --- & \$770 & Opaque \\
Qdrant Cloud (managed) & --- & \$1,116 & Opaque \\
\midrule
\textbf{Reduction} & --- & \textbf{3.1--9.0$\times$} & --- \\
\bottomrule
\multicolumn{4}{l}{\scriptsize $^\dagger$From Eq.~\ref{eq:memory}: codes ($10\text{M}{\times}1536{\times}b/8$) + norms ($10\text{M}{\times}4$) + IDs ($10\text{M}{\times}8$)} \\
\multicolumn{4}{l}{\scriptsize \phantom{$^\dagger$}+ block metadata ($10\text{M}{\times}0.5$) + rotation (9~MB shared). 4-bit:} \\
\multicolumn{4}{l}{\scriptsize \phantom{$^\dagger$}$7.68 + 0.04 + 0.08 + 0.005 + 0.009 = 7.81$~GB $= 7.27$~GiB $\approx 7.3$~GiB.} \\
\multicolumn{4}{l}{\scriptsize \phantom{$^\dagger$}Cost estimates from public pricing, June 2026; do not equalize recall/latency/SLAs.} \\
\multicolumn{4}{l}{\scriptsize $^\ddagger$\textbf{Exact}: closed-form per-tenant from Eq.~\ref{eq:memory}; \textbf{Possible}: linear but no API;} \\
\multicolumn{4}{l}{\scriptsize \phantom{$^\ddagger$}\textbf{Opaque}: no per-tenant memory API exposed to consumers.}
\end{tabular}
\end{table}

The ``Attribution'' column highlights a key differentiator: managed vector databases do not, to our knowledge, provide APIs for per-tenant memory consumption reporting, making cost governance difficult at the retrieval layer. Self-hosted FP32 indexes can theoretically be attributed (memory is linear in vector count) but lack the compression benefits.

\textbf{Pricing assumptions.} TurboVec SPCS cost assumes a single \texttt{CPU\_X64\_S} node at Snowflake's published SPCS credit rate (us-west-2, June 2026). Pinecone estimate assumes the Serverless plan at 10M stored vectors with 1,000 reads/sec (no writes); Qdrant estimate assumes a dedicated cluster sized for 10M vectors at $d=1536$ in AWS us-east-1. All estimates exclude replication, backups, and availability SLAs. These figures are illustrative scenarios, not rigorous iso-quality benchmarks.

\subsection{SPCS Deployment Performance}

\begin{table}[htbp]
\caption{TurboVec SPCS Service Performance (100K Vectors, $d=1536$, 4-bit, \texttt{CPU\_X64\_S} Instance Family)}
\label{tab:spcs}
\centering
\begin{tabular}{lc}
\toprule
\textbf{Metric} & \textbf{Value} \\
\midrule
Ingest throughput & 19,966 vectors/sec \\
Search latency (unfiltered) & 13.0 ms (77 QPS) \\
Search latency (10-tenant filtered) & 8.0 ms (125 QPS) \\
Recall@5 vs exact FP32 & 0.962 \\
Telemetry overhead & 0.005 ms (0.04\%) \\
Index memory (codes only) & 76.8 MB \\
Total serving memory (Eq.~\ref{eq:memory}) & 78.1 MB \\
Compression ratio vs FP32 & 7.8$\times$ \\
\bottomrule
\end{tabular}
\end{table}

Filtered search is \emph{faster} than unfiltered because the kernel short-circuits SIMD blocks with no allowed tenant vectors. Telemetry record creation adds 0.005~ms per query---negligible relative to search latency---enabling per-query cost attribution without measurable performance degradation.

\subsection{Comparison Against Alternative Index Architectures}

To contextualize TurboVec's governance advantage, we benchmark alternative FAISS index types at 999K vectors ($d=1536$) from the same DBpedia dataset:

\begin{table}[htbp]
\caption{Index Architecture Comparison at 999K Vectors: Recall, Latency, Memory, and Cost Attribution Properties}
\label{tab:index_comparison}
\centering
\begin{tabular}{lccccl}
\toprule
\textbf{Method} & \textbf{R@5} & \textbf{ms/q} & \textbf{Mem} & \textbf{Train} & \textbf{Attrib.} \\
\midrule
TurboQuant 4-bit & 0.968 & 17.1 & 767 MB & None & Exact \\
TurboQuant 2-bit & 0.901 & 9.1 & 384 MB & None & Exact \\
FAISS PQ 4-bit & 0.883 & 154 & 384 MB & 61s & Ambiguous \\
FAISS PQ 8-bit & 0.961 & 222 & 767 MB & 132s & Ambiguous \\
FAISS IVF-PQ & 0.840 & 5.7 & 390 MB & 35s & Ambiguous \\
HNSW-Flat$^*$ & 0.984 & 0.5 & 3200 MB & 758s & Impossible \\
\bottomrule
\multicolumn{6}{l}{\scriptsize $^*$HNSW measured at 500K (infeasible at 999K on 16~GB test machine).}
\end{tabular}
\end{table}

HNSW achieves the highest recall (0.984) but its graph-based memory is non-decomposable per tenant---edges cross tenant boundaries, making exact cost attribution structurally impossible without physical partitioning. We note that the HNSW comparison is at 500K rather than 999K due to hardware memory constraints ($>$6.4~GiB required for HNSW-Flat at 999K on our 16~GiB test machine); at 999K, HNSW memory would be approximately 6.4~GiB, further widening the memory gap versus TurboVec's 767~MB. IVF-PQ and flat PQ share trained codebook state across tenants, creating attribution ambiguity for the codebook memory component. Only TurboVec's codebook-oblivious, flat-scan architecture supports exact per-tenant memory decomposition via~\eqref{eq:memory}.

\subsection{Tenant Scaling Analysis}

To understand cost attribution behavior as tenant count varies, we measure attribution accuracy across $T \in \{10, 50, 100, 500, 1000\}$ tenants (10M vectors, fixed total):

\begin{table}[htbp]
\caption{Cost Attribution Accuracy vs. Tenant Count (10M Vectors Total, TurboVec 4-bit)}
\label{tab:tenant_scaling}
\centering
\begin{tabular}{lccc}
\toprule
\textbf{Tenants} & \textbf{Attribution} & \textbf{Max Error} & \textbf{Avg Vecs/Tenant} \\
\midrule
10 & 99.99\% & 0.1\% & 1,000,000 \\
50 & 99.95\% & 0.5\% & 200,000 \\
100 & 99.88\% & 1.2\% & 100,000 \\
500 & 99.72\% & 2.8\% & 20,000 \\
1000 & 99.53\% & 4.7\% & 10,000 \\
\bottomrule
\end{tabular}
\end{table}

Attribution accuracy degrades gracefully as tenant count increases because the fixed overhead ($O_{\text{fixed}}$) is amortized over more tenants with smaller per-tenant allocations. Even at 1000 tenants, attribution remains above 99.5\%---well within acceptable FinOps tolerance~\cite{finops2024}.

\subsection{Cost Breakdown Analysis}

We analyze the relative contribution of each cost layer across tenant sizes:

\begin{table}[htbp]
\caption{Cost Layer Breakdown by Tenant Size (1000 QPS Total, Claude 3.5 Sonnet Generation)}
\label{tab:breakdown}
\centering
\begin{tabular}{lccc}
\toprule
\textbf{Tenant Size} & \textbf{Embed \%} & \textbf{Retrieve \%} & \textbf{Generate \%} \\
\midrule
Small (10K vecs) & 8\% & 5\% & 87\% \\
Medium (100K vecs) & 8\% & 12\% & 80\% \\
Large (500K vecs) & 8\% & 28\% & 64\% \\
\bottomrule
\end{tabular}
\end{table}

For large tenants, retrieval constitutes up to 28\% of total RAG cost---far from negligible. Without retrieval-layer attribution, these tenants would be cross-subsidized by smaller tenants under flat-rate pricing.

\section{Discussion}
\label{sec:discussion}

\textbf{Why cloud-native deployment matters for governance.} External vector databases (Pinecone, Qdrant, Weaviate) cannot export per-tenant memory utilization back to the enterprise governance layer. By deploying TurboVec within SPCS, retrieval telemetry writes to the same Snowflake tables as generation telemetry---enabling unified SQL-based cost dashboards, anomaly alerts, and automated chargeback without cross-platform data movement or API polling.

\textbf{Codebook-oblivious quantization as a governance enabler.} The connection between codebook design and cost governance is under-explored. TurboVec's codebook contains no corpus-dependent state, which simultaneously: (a) removes the shared-codebook leakage surface present in trained PQ codebooks~\cite{shukla2026turbovec}, and (b) makes per-tenant memory deterministically attributable (no shared learned state to amortize). This dual benefit suggests that leakage-resistant system design can be synergistic with---rather than in tension with---governance requirements. We note that this observation is limited to the codebook surface; it does not constitute end-to-end privacy guarantees, which would require a formal threat model and empirical evaluation beyond the scope of this paper.

\textbf{Sustainability implications.} At enterprise scale (100M+ vectors), the 8$\times$ memory reduction from 4-bit quantization translates directly to fewer compute nodes. Concretely: 100M vectors at FP32 require $\sim$572~GiB of index memory, necessitating $\sim$5 r6i.4xlarge instances (128~GiB RAM each, at 90\% utilization). At 4-bit, the same corpus fits in $\sim$112~GiB, requiring $\sim$1 instance---a reduction of 4 nodes. At $\sim$0.25~kW per instance and US average grid carbon intensity of 0.39~kg CO$_2$/kWh~\cite{epa_egrid2024}, the annual savings are approximately $4 \times 0.25 \times 8760 \times 0.39 \approx 3.4$ tonnes CO$_2$ per 100M-vector deployment---aligning enterprise RAG infrastructure with sustainability goals~\cite{green_ai2020}.

\textbf{Integration with FinOps frameworks.} The three-layer cost model maps directly to the FinOps Foundation's FOCUS specification~\cite{finops2024}: each telemetry record contains a resource identifier (tenant\_id), usage quantity (vectors\_scanned, tokens), and unit rate---enabling standard FinOps tooling (CloudHealth, Apptio) to process RAG costs alongside traditional cloud spend.

\section{Related Work}
\label{sec:related}

\textbf{LLM cost optimization.} UCCI~\cite{ucci2026} routes queries to cheaper models via cascade policies. ProCut~\cite{procut2025} compresses prompts to reduce input token costs. FrugalGPT~\cite{frugalgpt2023} combines model selection, caching, and query adaptation. These operate exclusively at the generation layer; to our knowledge, none address retrieval cost attribution.

\textbf{RAG system optimization.} RAG-Stack~\cite{ragstack2025} co-optimizes retrieval quality and throughput but does not formalize cost per tenant. RAGO~\cite{rago2025} provides systematic RAG serving performance optimization. HyperRAG~\cite{hyperrag2025} addresses distributed retrieval scaling. To our knowledge, none provide per-tenant cost observability.

\textbf{Vector quantization for ANN.} Product Quantization~\cite{faiss} and OPQ~\cite{opq2013} achieve high compression via corpus-dependent training. TurboQuant~\cite{turboquant2026} derives boundaries analytically without training. RaBitQ~\cite{rabitq2024} provides theoretical error bounds. Our contribution is connecting quantization design choices to cost governance properties.

\textbf{Multi-tenant AI systems.} OptiLeak~\cite{optileak2026} demonstrates prompt reconstruction risks in shared LLM KV-caches. Multi-tenant database isolation is well-studied~\cite{multitenant_db2013}, but multi-tenant vector index isolation for both leakage reduction and cost attribution is, to our knowledge, novel.

\textbf{MLOps and AI observability.} Langfuse~\cite{langfuse2024} and LangSmith~\cite{langsmith2024} provide LLM tracing. MLflow~\cite{mlflow2018} tracks ML experiments. Helicone~\cite{helicone2025} provides LLM cost analytics. To our knowledge, none integrate retrieval-layer cost attribution into the observability pipeline.

\textbf{FinOps for AI.} The FinOps Foundation's FOCUS specification~\cite{finops2024} standardizes cloud cost data but has no provisions for AI-specific cost components (embeddings, retrieval, generation). Our three-layer model extends FinOps principles to the RAG stack.

\section{Limitations and Future Work}
\label{sec:limitations}

\begin{enumerate}
    \item \textbf{Simulated workload only.} The 100-tenant evaluation uses synthetic log-normal tenant distributions and proportional query loads. Real enterprise workloads exhibit bursty, time-varying patterns that may stress the attribution model differently. Validation against anonymized production traces from a multi-tenant RAG deployment is planned as immediate future work.
    \item \textbf{CPU flat-scan only.} TurboVec currently uses flat-scan search. Graph-based indexes (HNSW) have non-linear memory profiles that would require a different attribution formula. Extending the cost model to graph-augmented TurboVec is future work.
    \item \textbf{Single pricing snapshot.} Cost comparisons use 2026 public pricing that may change. The architecture is pricing-model-agnostic; only the rate constants ($r_{\text{mem}}$, $r_{\text{cpu}}$, $r_{\text{model}}$) require updating.
    \item \textbf{No dynamic rebalancing.} The current model assumes static tenant-to-vector assignments. Tenant growth, churn, and data migration would require periodic reconciliation of attribution baselines.
    \item \textbf{Single embedding model.} All cost projections assume $d=1536$ (OpenAI text-embedding-3-large with the \texttt{dimensions=1536} parameter; the model's native output is $d=3072$). Lower-dimensional embeddings would reduce per-tenant memory proportionally, potentially shifting the cost balance toward generation-dominated workloads.
\end{enumerate}

\textbf{Future Work.} Natural extensions include: (a) integration with graph-based search layers and corresponding non-linear attribution models; (b) real production deployment with actual chargeback validation; (c) cost-aware query routing that dynamically selects bit-width based on tenant SLA tier; and (d) extending the model to multimodal RAG (image + text embeddings with heterogeneous dimensionality).

\section{Conclusion}
\label{sec:conclusion}

Cost-Governed RAG demonstrates that codebook-oblivious vector quantization simultaneously removes the shared-codebook leakage surface and enables deterministic per-tenant cost attribution---bridging two communities (retrieval systems and AI governance) that have operated largely independently despite serving the same enterprise deployments. Deployed on a cloud data platform's container service, the architecture achieves 99.96\% end-to-end cost attribution accuracy while reducing retrieval infrastructure cost by 3.1--9.0$\times$ compared to managed alternatives under the pricing assumptions described. For large tenants, retrieval constitutes up to 28\% of total RAG cost---a governance blind spot that our unified observability stack makes visible. As enterprise AI moves toward production-grade multi-tenant deployment, cost governance must evolve from token-level metering to full-stack attribution spanning every layer of the RAG pipeline.

\section*{Acknowledgment}
Navnit Shukla conceived the Cost-Governed RAG architecture, designed the three-layer cost model, conducted all deployment experiments and cost analysis, and authored the complete manuscript.

\end{document}